\documentclass[letterpaper]{article} 
\usepackage{aaai25}  
\usepackage{times}  
\usepackage{helvet}  
\usepackage{courier}  
\usepackage[hyphens]{url}  
\usepackage{graphicx} 
\usepackage{natbib}  
\usepackage{caption} 
\usepackage{algorithm}
\usepackage{algorithmic}
\usepackage{makecell}
\usepackage{amsmath}
\usepackage{amssymb}
\usepackage{subfigure}
\DeclareMathOperator*{\argmax}{arg\,max}

\usepackage{newfloat}
\usepackage{listings}
\DeclareCaptionStyle{ruled}{labelfont=normalfont,labelsep=colon,strut=off} 
\floatstyle{ruled}
\newfloat{listing}{tb}{lst}{}
\floatname{listing}{Listing}
\title{Data Augmentation for Instruction Following Policies via Trajectory Segmentation}
\author {
    Niklas Hoepner\textsuperscript{\rm 1},
    Ilaria Tiddi\textsuperscript{\rm 2},
    Herke van Hoof\textsuperscript{\rm 1}
}
\affiliations {
    \textsuperscript{\rm 1}University of Amsterdam\\
    \textsuperscript{\rm 2}Vrije Universiteit Amsterdam\\
    n.r.hopner@uva.nl, i.tiddi@vu.nl, h.c.vanhoof@uva.nl
}

\usepackage{bibentry}

\begin{document}

\maketitle

\begin{abstract}
The scalability of instructable agents in robotics or gaming is often hindered by limited data that pairs instructions with agent trajectories. However, large datasets of unannotated trajectories containing sequences of various agent behaviour (play trajectories) are often available. In a semi-supervised setup, we explore methods to extract labelled segments from play trajectories. The goal is to augment a small annotated dataset of instruction-trajectory pairs to improve the performance of an instruction-following policy trained downstream via imitation learning. Assuming little variation in segment length, recent video segmentation methods can effectively extract labelled segments. To address the constraint of segment length, we propose Play Segmentation (PS), a probabilistic model that finds maximum likely segmentations of extended subsegments, while only being trained on individual instruction segments. Our results in a game environment and a simulated robotic gripper setting underscore the importance of segmentation; randomly sampled segments diminish performance, while incorporating labelled segments from PS improves policy performance to the level of a policy trained on twice the amount of labelled data.
\end{abstract}

\begin{links}
\link{Code}{https://github.com/NikeHop/PlaySegmentation-AAAI2025}
\end{links}

\section{Introduction}

\begin{figure*}[h!]
    \centering
    \includegraphics[width=\linewidth]{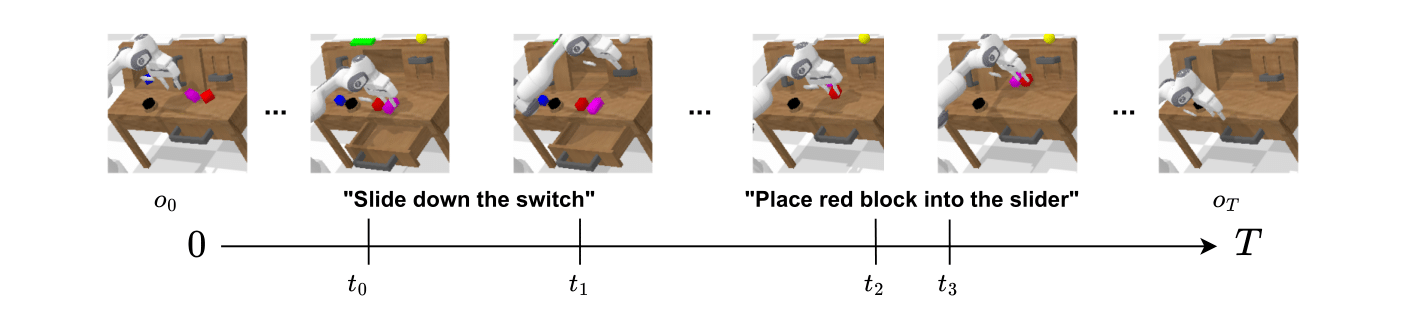}
    \caption{Example of play data, where the play trajectory contains sequences of instructable agent behaviour. The trajectory is represented by the observation sequence of the agent. Parts of the trajectory are labelled with the corresponding instructions and form the annotated dataset. Sampling random segments bears the risk to capture incomplete instructions or multiple ones.}
    \label{fig:play_data}
\end{figure*}

Advances in natural language conditioned generative models led to breakthrough results on text-to-image \cite{imagen}, text-to-text \cite{open_ai_instruction_following} and text-to-audio generation \cite{voicebox} forming the backbone of many recent AI applications \cite{gpt4}. A similarly capable model that learns to map natural language instructions to the corresponding trajectories of a game or robotic agent is an ongoing research challenge \cite{rtx2}. The main obstacle towards that goal is the lack of data, i.e. annotated trajectories containing diverse behaviours in different environments \cite{large_robot_dataset}.
To address this data limitation, recent work focuses on enhancing the sample efficiency of training algorithms \cite{sample_efficient_instruction_following}, transforming foundation models into policies \cite{rtx2} or performing data augmentation via simulation-based approaches \cite{DataSimulation}.
Another remedy for the lack of data is to develop training algorithms that learn from data sources that, although abundant, deviate from the conventional framework of instruction-trajectory pairs \cite{VIP,mimic_play}.  
One such source of less conventional data are large collections of videos of agent behaviour. For example, consider gameplay videos on YouTube \cite{VPTOpenAI} or play data in robotics \cite{original_play_data,calvin,mimic_play}, where long unsegmented trajectories are generated by humans teleoperating a robotic agent with the instruction to play. We will refer to trajectories that contain sequences of diverse agent behaviour as play trajectories. Our goal is to generate training data for an instruction following policy from them. This task requires segmenting the play trajectories and labelling the resulting segments with the corresponding instructions. We tackle this problem in a semi-supervised setting, in which a small amount of subsegments of the play trajectories are labelled with the corresponding instructions (see Figure \ref{fig:play_data}).

Previous work on trajectory segmentation has focused on unsupervised methods, aiming to discover skills that maximise the likelihood of the unsegmented trajectories \cite{compile, mixture_policies, love}. These approaches do not guarantee that the identified skills align with any natural language instructions. Research on labelling trajectory segments with instructions has generally assumed that labelled and unlabelled segments are drawn from the same distribution \cite{labelling_model,relabelling}. Fewer studies explore the segmentation of trajectories containing multiple behaviours into segments corresponding to individual instructions \cite{taco,sl3}. However, they assume that entire trajectories are labelled with natural language plans that outline the sequence of instructions executed by the agent.  

Extracting labelled segments from play trajectories poses
the challenge of identifying segments that align with specific instructions. Adding non-representative or wrongly labelled segments, potentially harms the performance of a policy trained downstream. We adapt existing
segmentation methods from the video segmentation literature and show that they struggle to generalise from the labelled dataset to the longer play trajectories. To address this, we introduce play segmentation (PS), a probabilistic model of segmentations of play trajectories, that can be trained from single instruction segments. To summarise, our contributions are:

\begin{itemize}
\item Adapting video segmentation methods to play data and analysing their ability to improve policy performance via data augmentation.
\item Developing Play Segmentation, a segmentation method capable of segmenting trajectories containing sequences of instructions while only being trained on individual instruction segments.
\item Demonstrating that the extracted labelled segments of Play Segmentation lead to policy improvements across two environments.
\end{itemize}

\section{Related Work}

\textbf{Trajectory Segmentation:} A significant body of work has focused on learning reusable skills from a set of trajectories in an unsupervised manner \cite{niekum_unsupervised,DDO,compile,love,llm_segment}. The underlying approach shared by these methods is to model the distribution of trajectories using a latent variable model, where the latent variables represents the active skill at each timestep. After training the probabilistic model via maximum likelihood \cite{DDO,compile}, inferring the latents for a trajectory results in a segmentation of the trajectory \cite{love}. However, there is no guarantee that the segments will correspond to specific instructions. To ensure that the extracted subsegments align with natural language instructions, a set of annotated examples can be employed \cite{taco}. Given the high cost of annotations, it is desirable for this set to be only a small fraction of the available demonstrations, which naturally leads to a semi-supervised setup \cite{sl3}. Prior work looked at data setups in which trajectories are paired with plans describing the sequence of skills performed by the agent \cite{taco,sl3}. Here, the annotation are segments of individual instructions within longer trajectories. Therefore we do not need to model the segmentation as a latent variable but can learn the parameters of a conditional distribution over segmentations given a trajectory.

\textbf{Semi-Supervised Instruction Following}: The standard approach to train instructable agents is to perform imitation learning on a dataset of instruction-trajectory pairs \cite{robotic_imitation_language,bc_z,interactive_language}. Due to the difficulty of generating natural language annotated trajectories, the problem is often studied in a semi-supervised learning setup \cite{labelling_model, relabelling}. For example, \citet{labelling_model} train a labelling  model using CLIP embeddings \cite{clip} on a small annotated dataset to then label a large unannotated dataset. Training a policy on the joint dataset improves the task accuracy. However, the unlabelled and labelled trajectories come from the same distribution, making it unnecessary to identify segments corresponding to instructions in the unlabelled trajectories. The model that has been trained in a setting closest to ours is SL3 \cite{sl3}. Given a small set of trajectories labelled with an overall goal and the sequence of instructions contained in it \cite{alfred}, SL3 applies an iterative procedure of segmenting trajectories, labelling segments and learning an instruction conditioned policy. The labelled dataset here consists of individual instructions and not plans, but unlike the case of SL3 does contain information about the start and end of these instructions.  

\textbf{Video Segmentation}: The vision community has studied the task of identifying actions in uncropped videos under related but distinct setups. While Action Segmentation (AS) methods \cite{st_cnn,ms_tcn, asformer,as_survey} try to predict for each frame the correct action class, Temporal Action Localization (TAL) methods try to predict the boundary timesteps of an action segment and a corresponding action label \cite{tallformer,tridet, tal_survey}. Some frameworks are developed to handle multiple of these tasks \cite{unloc}. Datasets for training contain uncropped videos containing multiple, not necessarily contiguous, action segments that are annotated on a frame level or with segment boundary information \cite{coin_dataset,assembly101_dataset,breakfast_dataset}. Both classes of methods have been studied under different levels of supervision \cite{unsupervised_as,semi_supervised_as,weakly_supervised_as}.
Our setting provides a unique challenge where the length of the training videos and the number of activities in the training videos differ from the videos at test time. To the best of our knowledge we are the first to investigate the usage of video segmentation models for data augmentation in sequential decision making settings.

\section{Methodology}

The full pipeline of our approach starts with segmenting play trajectories, followed by labelling the extracted segments with instructions and lastly training a policy on the extracted labelled segments. Depending on the segmentation model, the segmentation and labelling can be done jointly. First, we describe the structure of play data and introduce notation, followed by a description how video segmentation models can be adapted to the play data setting. Finally, we introduce our play segmentation approach.

\subsection{Play Data}
Given is a small annotated dataset $D_{\textrm{ann}}=(\tau_{k},d^{T_{k}}_{k})^{N}_{k=1}$ of instruction-trajectory pairs, next to a large unannotated dataset of only play trajectories $D_{\textrm{unann}}=(d^{T^{'}_{k}}_{k})^{M}_{k=1}$, where $T_{k} \ll T^{'}_{k}$. Here, $d^{T}=(o_{0},a_{0},...,o_{T})$ denotes a trajectory of length $T+1$ with $o_{t}$ and $a_{t}$ being the observation and action of the agent at timestep $t$ and $\tau$ is an instruction in natural language from a finite set of possible instructions $\zeta$. We denote the segmentation of a trajectory as $\alpha_{0:T-1} \in \{0,1\}^{T}$, where $\alpha_{i}=1$ means that a segment ended at timestep $i$. A labelled segmented trajectory is the triple $(d^{T},\alpha_{0:T-1},\tau_{1:K})$, such that $\sum^{T-1}_{t=0} \alpha_{t} = K$.

\subsection{Segmentation Models} \label{subsec:seg_models}
Here, we discuss different strategies to extract labelled segments from play trajectories. We start with the simplest way, i.e. labelling random segments, and then discuss how video segmentation methods can improve upon this approach by cropping enlarged random segments. We then present Play Segmentation, a conditional probabilistic model over segmentations given play trajectories, that can be trained having access only to individual instruction segments.

\textbf{Random Segments}: The simplest method to augment the labelled dataset involves randomly sampling windows from the play trajectory and labelling them using a labelling model trained on $D_{\textrm{ann}}$. The segment length is sampled uniformly between the minimum and maximum length of the segments in $D_{\textrm{ann}}$. It is unlikely that random segments begin and end precisely at the start and end of an instruction, leading to out-of-distribution challenges for the labelling model. An alternative is to sample larger segments that contain the start and end of a single instruction and develop methods that crop out the instruction from the larger segment. We adapt UnLoc \cite{unloc}, an Action Segmentation method, and TriDet \cite{tridet}, a Temporal Action Localisation method, to perform the cropping. 

\begin{figure}[t!]
    \centering
    \includegraphics[width=\linewidth]{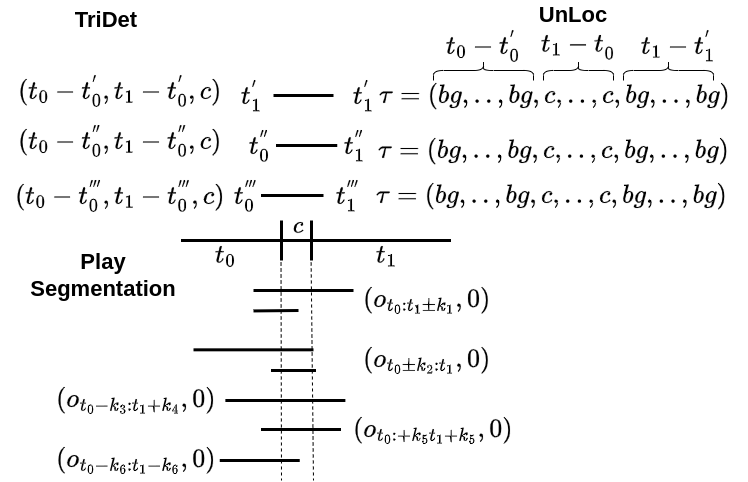}
    \caption{Overview of how training samples are generated from a single annotated segment for the different segmentation approaches. Here, $c$ stands for the instruction class of the segment and $bg$ for the background class.}
    \label{fig:neg_segments}
\end{figure}

\textbf{UnLoc \cite{unloc}}: UnLoc is a CLIP-based framework designed to tackle multiple video understanding tasks such as TAL and AS. Here, we focus on the action segmentation variation of UnLoc. It learns to predict frame-wise action labels given a segment of image-based environment observations by learning the probabilistic model $p_{\theta}(\tau_{0:T-1} | o_{0:T})=\prod_{t=0}^{T-1} p_{\theta}(\tau_{t}| o_{0:T})$. The image observations and natural language instructions are encoded via CLIP. For each instruction the embedding is concatenated to the end of the observation sequence. Then a self-attention-based architecture outputs the logit of the instruction label for each frame. The usual training data \cite{alfred} consists of pairs of segments and frame-wise annotations, containing multiple actions and possibly background segments. Since the longer play trajectories do not have frame-wise annotations, we approximate this setup by sampling windows around the annotated segments and labelling the additionally sampled frames as background (see Figure \ref{fig:neg_segments}). 

At test time the trained model can provide frame-wise labels for the whole play trajectory. However, during training the model only receives input segments containing one action-segment. We sample windows of the same length during training. The final labelled segment can then be obtained by cropping away the frames that are labelled as background. The details of the cropping procedure based on the individual frame predictions are explained in Appendix \ref{app:seg_models}. UnLoc does not scale with the number of instructions as each segment needs to be processed with each possible instruction such that the batch size is multiplied for each sample by the number of possible instructions $|\zeta|$. 

\textbf{Tridet \cite{tridet}}: TAL methods are commonly trained on a dataset consisting of uncropped video segments $o_{0:T}$ containing multiple action segments, where each action segment is given by a triplet of the form $(s_{i},e_{i},c_{i})$ \cite{HACS}. Here $s_{i}$ represents the starting timestep, $e_{i}$ the ending timestep and $c_{i}$ the action class.
The key design aspects of most architectures are pretrained features from an action classifier, followed by a feature pyramid to account for action segments of different length \cite{tal_survey}. The TriDet architecture proposes a new boundary prediction head on top of the feature pyramid achieving state of the art performance on multiple TAL benchmarks. Similarly to UnLoc, we sample random windows around the segments of the annotated dataset and train TriDet to identify the action segment in the enlarged segment. At inference time, we sample random windows from the play trajectories and let TriDet identify the action segment in the large random window. As TriDet directly predicts the boundary locations and action class, no further postprocessing is necessary before adding the labelled segments to the annotated dataset. For more details on TriDet we refer to Appendix \ref{app:seg_models}.

One drawback from the cropping approach is the assumption that there exist a window size for the random segment that captures with high likelihood the start and end of exactly one instruction. If the shortest and longest instruction length differ a lot, a large window size potentially captures multiple shorter instructions and a shorter window size will never capture a longer instruction. It is important to note that UnLoc and TriDet both are capable of segmenting full play trajectories, if fully segmented play trajectories were part of the training data.

\textbf{Play Segmentation (PS)}: We introduce Play Segmentation (PS), a model tailored to segmenting play data. The desideratum for the model is that it can be trained on short segments from $D_{\textrm{ann}}$, but generalises to segmenting longer play trajectories. We start by factorising the probabilistic model over labelled segmentations $(\tau_{1:K},\alpha_{0:T-1})$ as follows:

\begin{equation*}
\begin{split}
    p(\tau_{1:K},\alpha_{0:T-1}&|o_{0:T})  \\
    = p_{\theta_{\textrm{LM}}}(\tau_{1:K}|\alpha_{0:T-1},o_{0:T})&\cdot p_{\theta_{\textrm{Seg}}}(\alpha_{0:T-1}|o_{0:T}) \\
    = \prod^{K}_{k=1} p_{\theta_{\textrm{LM}}}(\tau_{k}|o_{\boldsymbol \alpha(k): \boldsymbol \alpha(k+1)}) 
     &\cdot\prod^{T-1}_{t=0} p_{\theta_{\textrm{Seg}}}(\alpha_{t} | o_{\gamma(\alpha_{0:t}):t+1}),
\end{split}
\end{equation*}

where $\boldsymbol \alpha$ maps the $i$-th segment to the timestep it starts, $\alpha(K+1)=T$ and $\gamma(\alpha_{0:t})=\max\{i|\alpha_{i}=1,i\in \{0,...,t\}\}$.
Next to a model that can predict the correct label for a segment ($p_{\theta_{\textrm{LM}}}$), we need to train a model to predict whether an observation sequence corresponds to an instruction segment ($p_{\theta_{\textrm{Seg}}}$). However the labelled dataset contains only segments that correspond to instructions. To train $p_{\theta_{\textrm{Seg}}}$, we augment $D_{\textrm{ann}}$ with different types of negative samples. 

For each segment of the annotated dataset $(\tau,o_{t_{0}:t_{1}})$ we translate the observation sequence $o_{t_{0}:t_{1}}$ to the left and right by a random number of timesteps sampled from $\{t_{\textrm{min}},...,t_{\textrm{max}}\}$ to form $D^{\textrm{aug}}_{\textrm{ann}}$. The minimum and maximum steps for the translation are hyperparameters of the method and should be chosen based on the segment lengths present in the annotated dataset. Additionally we obtain negative samples by sampling a random timestep from $\{t_{\textrm{min}},...,t_{\textrm{max}}\}$  used for elongating or shortening the groundtruth segment. For a single annotated segment $(o_{t_{0}:t_{1}},\alpha=1)$ we obtain the following negative segments:

\begin{itemize}

    \item $(o_{t_{0}:t_{1} \pm k_{1}},\alpha=0)$, $(o_{t_{0} - k_{2}:t_{1}},\alpha=0)$
    \item $(o_{t_{0}-k_{3}:t_{1}+k_{4}},\alpha=0)$,
    \item $(o_{t_{0}+k_{5}:t_{1}+k_{5}},\alpha=0)$,
    $(o_{t_{0}-k_{6}:t_{1}-k_{6}},\alpha=0)$.
\end{itemize}
The model is trained by minimising the loss:

\begin{equation*}
    \begin{split}
        L(\theta_{\textrm{LM}},\theta_{\textrm{Seg}}) &= \\
        \mathbb{E}_{s \sim D_{\textrm{ann}}}[\log(p_{\theta_{\textrm{LM}}}(\tau|o_{t_{0}:t_{1}})) &+ \log(p_{\theta_{\textrm{Seg}}}(1|o_{t_{0}:t_{1}})) ] \\ 
        + \mathbb{E}_{s^{'} \sim D^{\textrm{aug}}_{\textrm{ann}}}[\log(p_{\theta_{\textrm{Seg}}}(0|o_{t_{0}:t_{1}}))&],  \\
    \end{split}
\end{equation*}

where $s=(o_{t_{0}:t_{1}},1,\tau)$ and $s^{'}=(o_{t_{0}:t_{1}},0)$. The labelling model is only trained on the positive samples as the negative ones cannot be labelled with an instruction. Both models share parameters, but have separate prediction heads. At segmentation time we  determine the most likely segmentation $\alpha^{\ast}_{0:T-1}$ using $p_{\theta_{\textrm{Seg}}}$:

\begin{equation}\label{eq:segmentation_distribution}
\begin{split}
    & \alpha_{0:T-1}^{\ast} = \argmax_{\alpha_{0:T-1}} p_{\theta_{\textrm{Seg}}}(\alpha_{0:T-1}|o_{0:T}) \\ 
\end{split}
\end{equation}

We can use dynamic programming (DP) to find $\alpha^{\ast}_{0:T-1}$ via the recursion:

\begin{equation} \label{eq:recursion}
\begin{split}
    &\max_{\alpha_{0:T-1}} \log p_{\theta_{\textrm{Seg}}}(\alpha_{0:T-1}|o_{0:T}) = \\ 
    &\max_{i\in\{0,...,T-1\}} ( \max_{\alpha_{0:i}} \log p_{\theta_{\textrm{Seg}}}(\alpha_{0:i} | o_{0:i+1}) + \\
    & \log p_{\theta_{\textrm{Seg}}}(\alpha_{i+1:T-1}=(0,...,1)|o_{i+1:T})) .
\end{split}
\end{equation}

The DP algorithm has cubic complexity in the number of timesteps and is related to the Viterbi algorithm for Hidden Semi Markov Models \cite{hsmm}. Due to this complexity we cannot segment full play trajectories for large $T$  but instead segment windows of size $\omega$. The first window is sampled starting at $t=0$, the beginning of the play trajectory. Let $o_{t_{k}:\omega}$ be the last segment resulting from the segmentation process. Then the starting point for the next window is either $\omega$ if $p_{\theta_{Seg}}(\alpha=1| o_{t_{k}:\omega})>0.5$, or $t_{k}$ otherwise. In that manner complete play trajectories can be segmented. In Table \ref{tab:pro_con_seg_model} we give an overview of the computational complexities of the introduced segmentation models at segmentation time with respect to the trajectory length. More details on the segmentation algorithm of PS can be found in Appendix \ref{app:seg_models}. 

\begin{table}[t]
    \centering
    \begin{tabular}{l|c|c|c}
        Method & UnLoc & TriDet & PS \\
        \hline
         Comp. Complexity & $\mathbb{O}(|\zeta|)$ & $\mathbb{O}(1)$ & $\mathbb{O}(T^{3})$ \\
    \end{tabular}
    \caption{Comparison of the computational (comp.) complexity of segmentation models. It is measured by the number of neural functional evaluations (NFE) needed to segment one trajectory, depending on the length of the trajectory $T$ and the number of instructions $|\zeta|$. Note that for PS the segmentation results potentially in multiple annotated segments.}
    \label{tab:pro_con_seg_model}
\end{table}

\section{Experiments}
The goal of the evaluation is to assess the capability of the different segmentation models to extract labelled segments from the play trajectories that can be used for data augmentation in an imitation learning context. In Section 4.1 the two evaluation environments are introduced and in Section 4.2 the importance of segmentation for successful data augmentation is highlighted. In Section 4.3 we compare the downstream-policy performance resulting from the different segmentation models and investigate reasons for the performance differences. 

\subsection{Environments}

\begin{figure}[t!]
\centering
\begin{subfigure}
  \centering
  \includegraphics[width=0.48\linewidth]{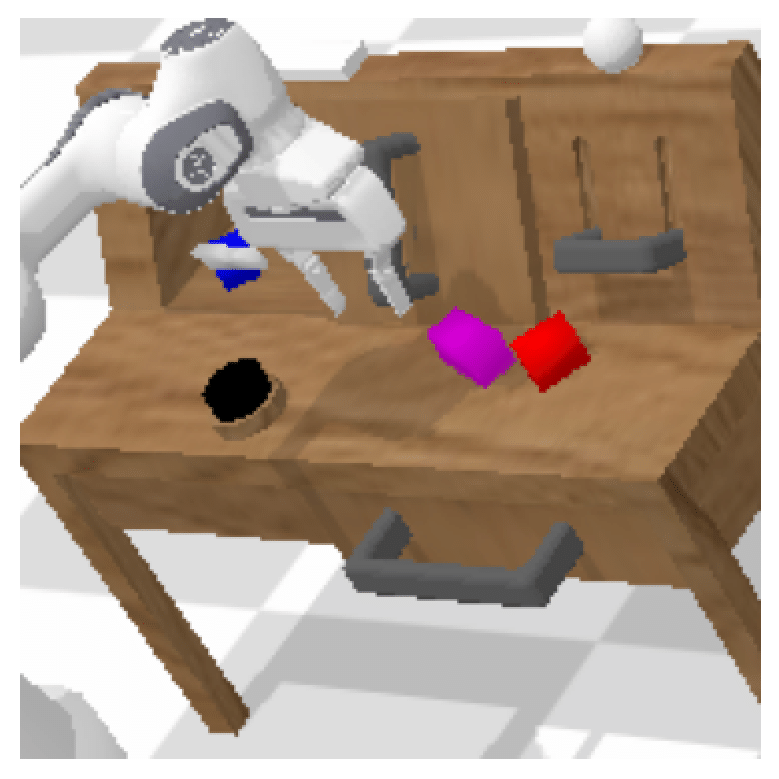}
  \label{fig:sub1}
\end{subfigure}%
\begin{subfigure}
  \centering
  \includegraphics[width=0.48\linewidth]{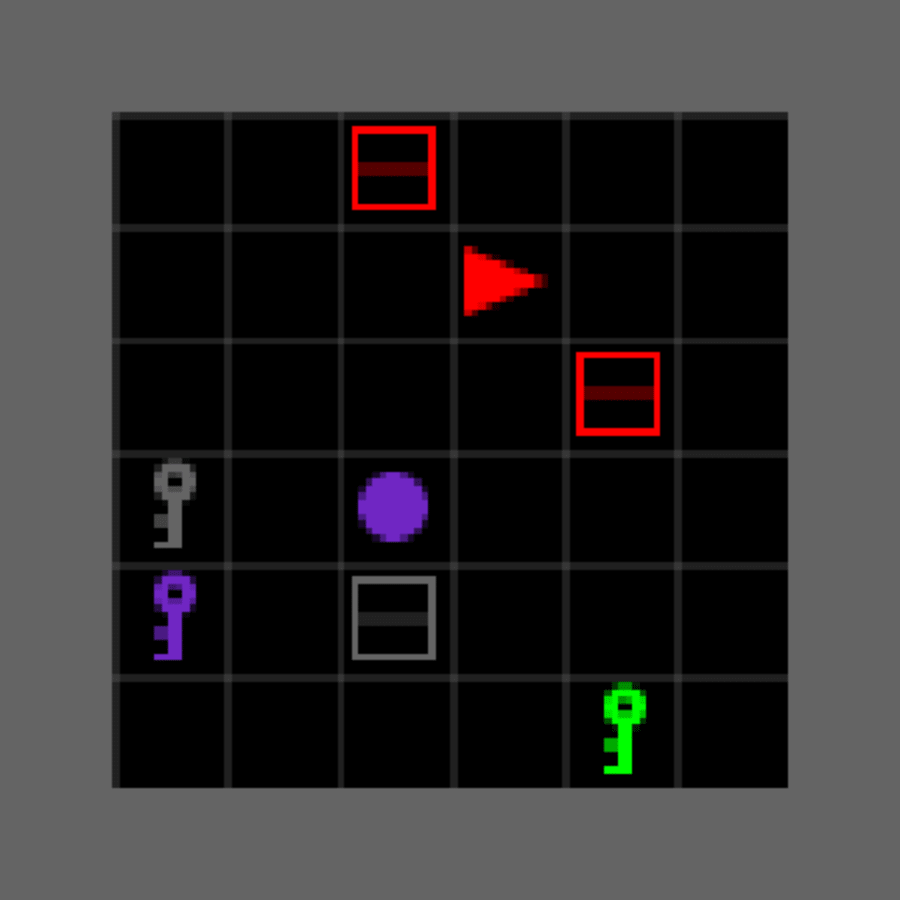}
  \label{fig:sub1}
\end{subfigure}%
\caption{Example observations from the CALVIN environment (left) and the BabyAI environment (right).}
\label{fig:datasets}
\end{figure}

\textbf{BabyAI} \cite{babyai} is a grid-based environment with a range of difficulty levels designed to test instruction following agents. Here we choose the GoTo-environment with 7 distractors (Figure \ref{fig:datasets}). The agent needs to navigate to the object described in the instruction. If multiple objects of the same type are present going to any of them solves the task. The objects can be one of three types with six different colours. The discrete action space is 4-dimensional and the observations are 64x64 RGB images. Demonstrations can be generated via a bot that solves the task. To generate the play trajectories we sample a new goal object every time the agent solves its current task until it has solved a sequence of 10 different tasks. The length of the unsegmented trajectories ranges from 11 to 71 timesteps. The controlled generation of the play trajectories allows us to evaluate the quality of the extracted labelled segments, as we know which instruction is active at each timestep. One limitation of the dataset is the possibility that at a single timestep two instructions are active, i.e. the agents goal is a red key but it comes across a yellow ball along the optimal path. We evaluate policies by measuring the percentage of tasks solved within 25 timesteps over 512 episodes. Given a dataset of instruction-trajectory pairs we train a policy via imitation learning following the proposed implementation by \citet{sample_efficiency_babyai}.
\noindent

\textbf{CALVIN} \cite{calvin} is a dataset containing play trajectories of a simulated 7-DOF Franka Emika Panda robot arm acting in a tabletop environment (Figure \ref{fig:datasets}). Tasks include manipulating objects of different colours and shapes as well as opening/closing doors and switching lights on/off. There exist $34$ different types of tasks. Individual segments of the play trajectories are labelled by human annotators with the corresponding instruction that is being executed. The length of the play trajectories varies between 1674 and 30838 timesteps, while the length of the instructions varies between 32 and 64 timesteps. The policy takes as input the image of a static camera showing the gripper arm and tabletop as well as an image of the gripper camera. An instruction following policy is evaluated on it capability of following sequences of natural language instructions. Each sequence consists of five instructions. The performance metric is the number of instructions completed by the policy until its first failure. The final evaluation score is the average number of instructions completed, computed over a 1000 sequences. Similar to prior work the policy is trained via multi-context imitation learning (MCIL) \cite{mcil_original,calvin,hulk}. Details on the policy training can be found in Appendix \ref{app:policy_training}.

\subsection{Does Segmentation Matter?}

\begin{figure}[t!]
\centering
\begin{subfigure}
  \centering
  \includegraphics[width=1\linewidth]{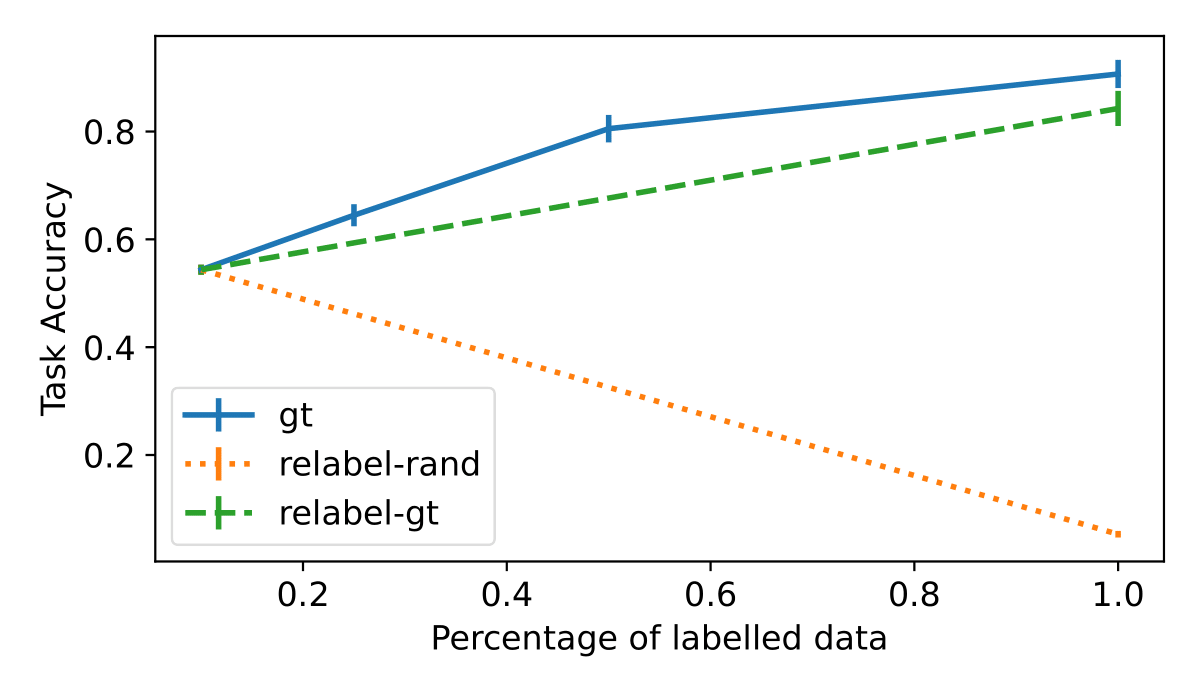}
  \label{fig:sub1}
\end{subfigure}%
\begin{subfigure}
  \centering
  \includegraphics[width=1\linewidth]{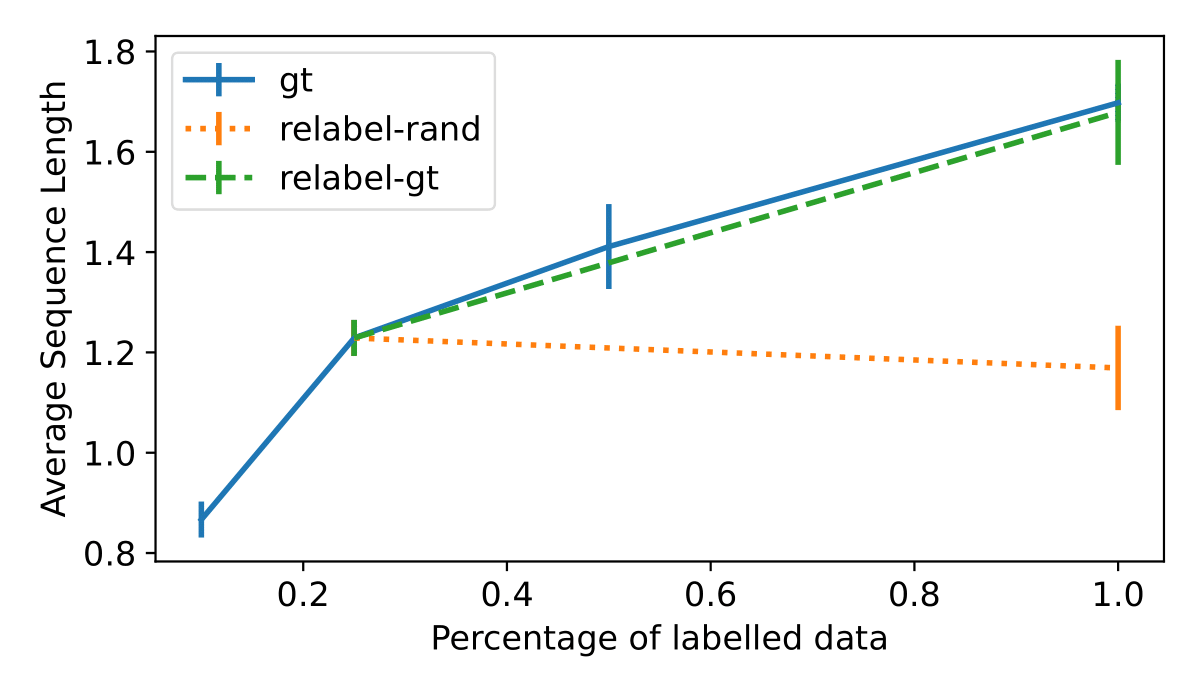}
  \label{fig:sub2}
\end{subfigure}
\caption{Effect of different amounts of annotated data on policy performance as well as the effect of data augmentation via labelled groundtruth segments and labelled random segment for BabyAI (top) and CALVIN (bottom).}
\label{fig:policy_performance}
\end{figure}

To assess the policy's performance with varying levels of annotated data, we subsample for both environments the annotated dataset into subsets of $10\%$, $25\%$, and $50\%$. Subsequently, the policy is trained for each subset. As anticipated, instruction completion decreases with less annotated data, as shown in Figure \ref{fig:policy_performance} (gt condition).
For the next steps in the analysis, we will attempt to recover the policy's performance at a $100\%$ of the labelled data by starting with a subset of it and adding labelled segments extracted from the unsegmented trajectories of the unlabelled dataset. For the BabyAI environment we start with a subset consisting of $10\%$ of the labelled dataset and for the CALVIN environment with a subset consisting of $25\%$ of the labelled dataset. In the following we will refer to these subsets as the starting split. 
To understand whether it matters which segments are labelled and used for data augmentation we compare the performance between adding randomly sampled segments and groundtruth segments to the starting split. First, we train a labelling model based on the I3D architecture from the action recognition literature \cite{i3d} on the starting split. More information on the training of the labelling model is in Appendix \ref{app:labelling_model}. It achieves a validation accuracy of $\sim 92\%$  BabyAI and $\sim 95\%$ on CALVIN. To add labelled groundtruth segments, we remove the labels from the left out data not in the starting split and add the predictions of our labelling model as new labels.

In Figure \ref{fig:policy_performance}, we can see that for both environments adding labelled groundtruth segments recovers the performance of a policy trained on all of the labelled data. One can conclude that the high accuracy transfers to the left out data only introducing a few wrongly labelled segments not having much impact on the policies performance. For the randomly sampled segments performance is either hampered (BabyAI) or stays the same (CALVIN). Adding randomly sampled segments has a larger negative effect in the BabyAI environment due to the greater variance in instruction length. This variance makes it less likely that a randomly sampled segment corresponds to a single completed instruction that can be correctly labelled by the labelling model.

\subsection{Comparison of Segmentation Models}

All segmentation models are trained on the starting split and then applied to extract labelled segments from the play trajectories. The starting split is augmented with labelled segments until it has the same number of instruction-trajectory pairs as the original labelled dataset. A policy is trained on top of the augmented dataset. We do not have any annotations for the play trajectories of the CALVIN dataset. This is a more realistic scenario but makes it harder to analyse the quality of the extracted labelled segments. Therefore, we start with the analysis in the BabyAI environment and then move to the CALVIN environment.

We measure the quality of the labelled segmentation via the accuracy of the assigned labels as well as the precision and recall of the segmentation points. It is important to note that these metrics are lower bounds for the actual performance. In the unsegmented trajectories of the BabyAI environment two instructions can overlap, i.e. while following instruction A the agent also completes instruction B. Therefore, the segmentation model can choose a correct segmentation point and label that is not part of the annotations. Video segmentation models have low precision and label accuracy (see Table \ref{tab:segmentation_quality}), indicating that the chosen segments do not correspond to groundtruth segments. This indicates that video segmentation models cannot generalise their segmentation performance from the training segments to the random segments from play trajectories.

\begin{figure}[t!]
\centering
\begin{subfigure}
  \centering
  \includegraphics[width=0.48\linewidth]{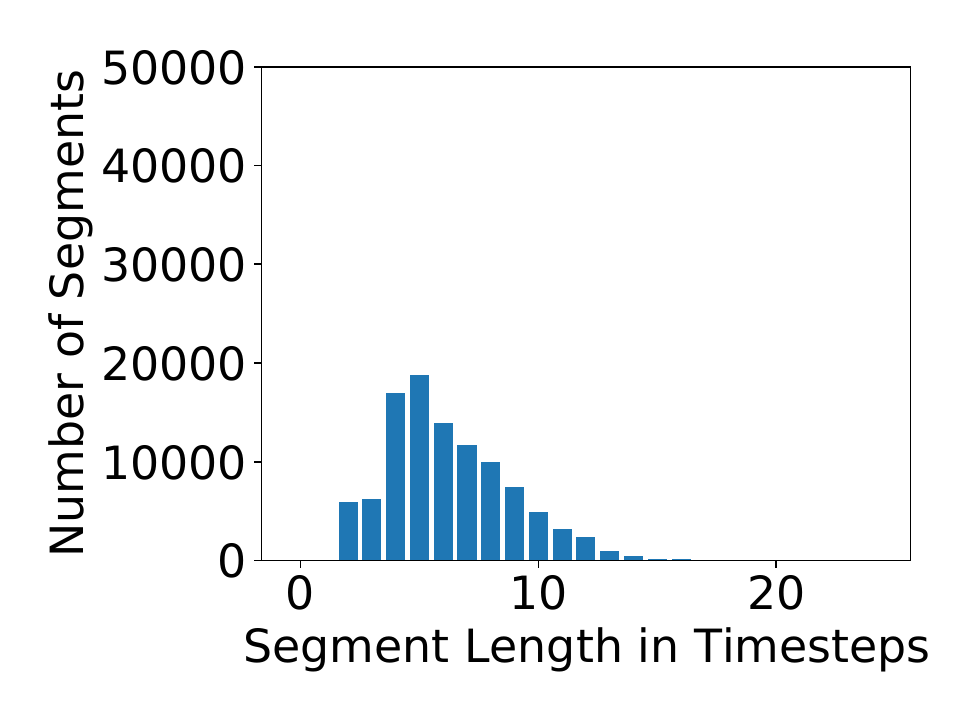}
\end{subfigure}%
\begin{subfigure}
  \centering
  \includegraphics[width=0.48\linewidth]{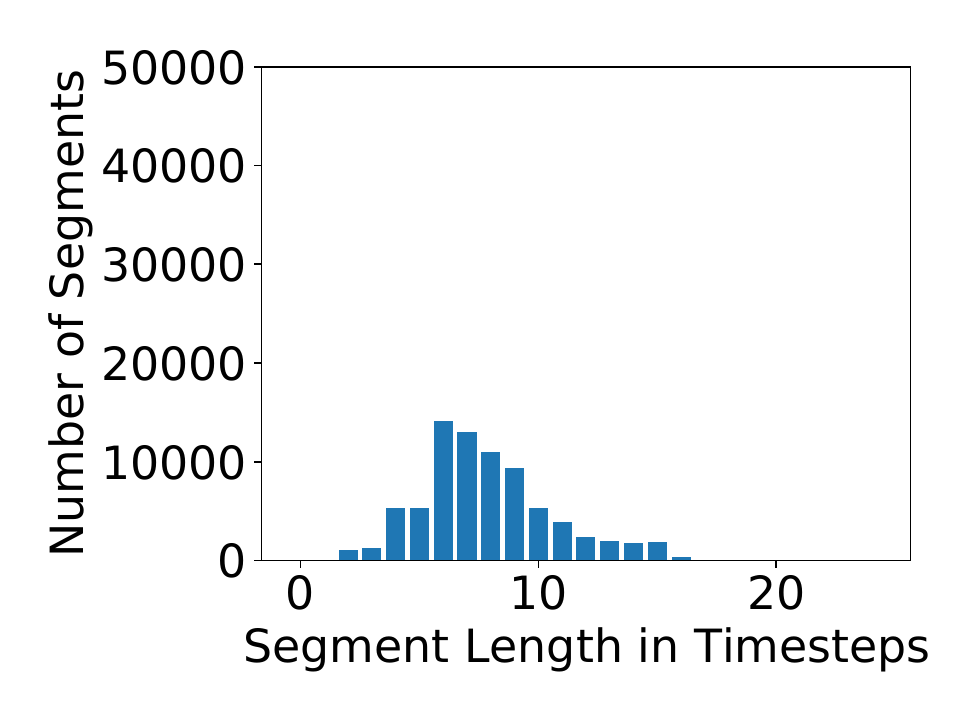} 
\end{subfigure}%
\begin{subfigure}
  \centering
  \includegraphics[width=0.48\linewidth]{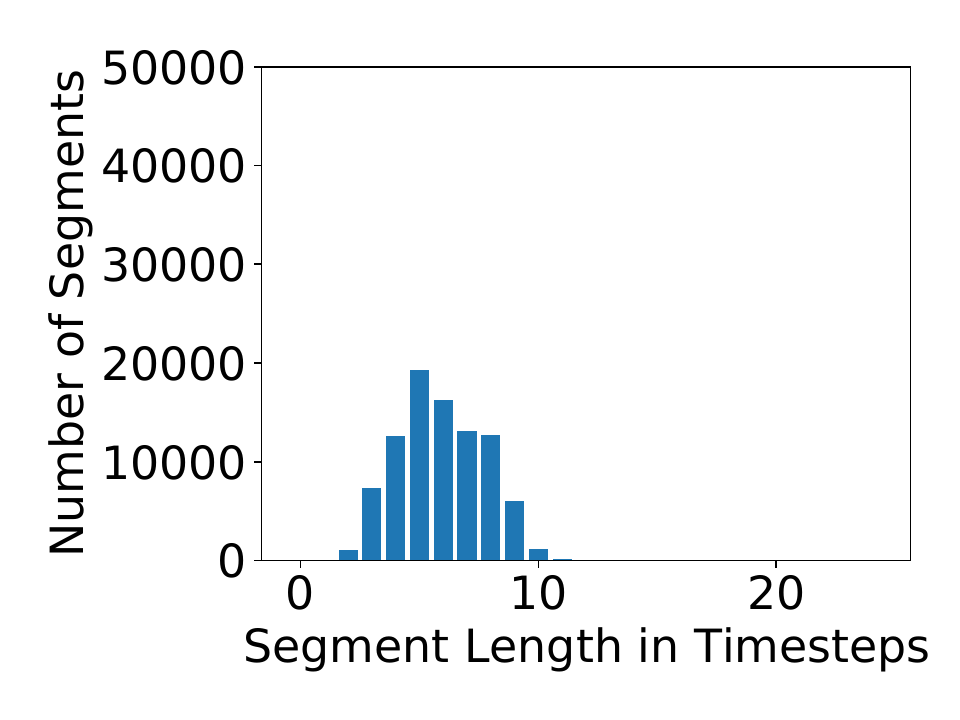}
\end{subfigure}%
\begin{subfigure}
  \centering
  \includegraphics[width=0.48\linewidth]{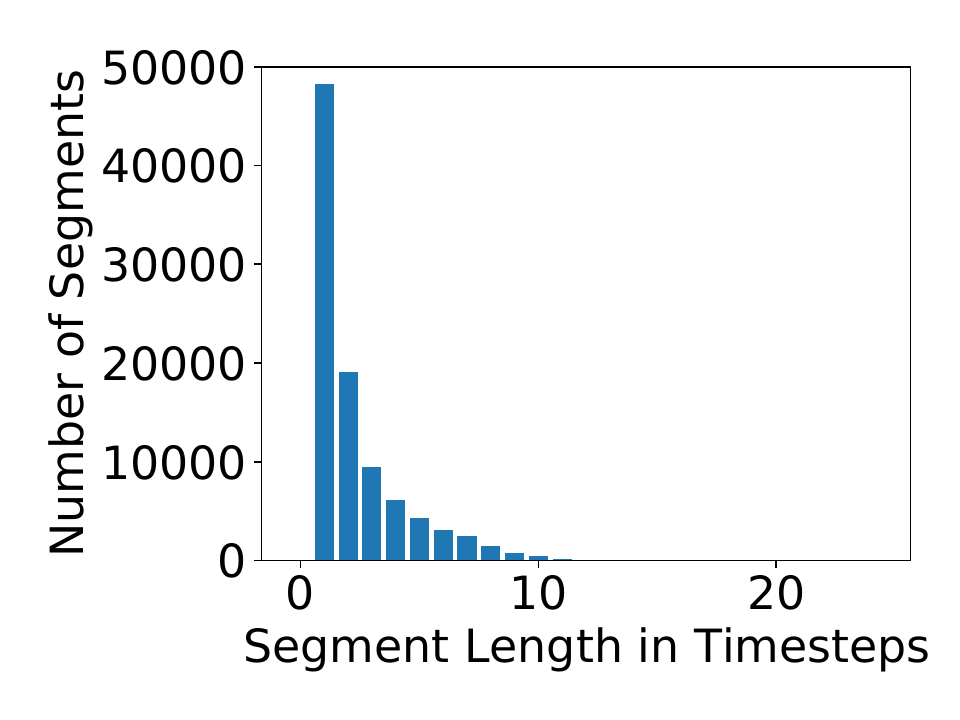}
\end{subfigure}
\caption{Distribution of segment length of the segments present in the groundtruth data (top left) and extracted via Play Segmentation (top right), TriDet (bottom left) and UnLoc (bottom right).}
\label{fig:segment_length_distribution}
\end{figure}

Play Segmentation achieves a higher precision and label accuracy than both video segmentation models. The recall of play segmentation is notably lower than the precision, indicating undersegmentation. This is also visible from the distribution of segment length of the extracted segments (Figure \ref{fig:segment_length_distribution}). While the segments obtained via play segmentation have a similar length distribution compared to the groundtruth, the number of segments with 1-3 timesteps is smaller. The performances of the policies trained downstream on the augmented datasets reflect the segmentation performances. While extracted labelled segments from TriDet have a negative impact on policy performance, data augmentation via Play Segmentation improves the performance to a level higher than if we had twice the amount of labelled data available (see Table \ref{tab:augmentation_results_babyai}). The extracted segments from UnLoc have little effect on the policies performance. This is a result of the short segment length of the extracted segments (Figure \ref{fig:segment_length_distribution}). During training of the policy a state of the trajectory is sampled jointly with the corresponding instruction to predict the next action. As a result longer segments make up a larger part of the policy training dataset.

\begin{table}[t]
    \centering
    \begin{tabular}{l|c|c|c|c}
    Method & Precision & Recall & F1 & Label Acc.  \\
    \hline
    PS  &  0.826 & 0.627 & 0.716 & 0.516\\
    UnLoc  & 0.283  & - & - & 0.353 \\
    TriDet  & 0.338 & - & - & 0.323 \\
    \end{tabular}
    \caption{Quality of the extracted labelled segments. For UnLoc and TriDet we do not measure recall as they do not provide a complete segmentation of a trajectory but only the start and end of a single segment.}
    \label{tab:segmentation_quality}
\end{table}

\begin{table}[t]
    \centering
    \begin{tabular}{l | c }
    \hline
    Dataset & Task Accuracy ($\pm$ Std.) \\
    \hline
    GT - 100\% & 0.907 $\pm$ 0.026 \\
    GT - Relabel & 0.843 $\pm$ 0.032 \\
    GT - 50 \% & 0.805 $\pm$ 0.025 \\
    PS &   \textbf{0.702 $\pm$   0.023} \\
    GT - 25\% & 0.645 $\pm$ 0.020 \\
    UnLoc &  0.583 $\pm$  0.020   \\ 
    GT - 10\%  & 0.544 $\pm$ 0.009 \\
    TriDet & 0.261 $\pm$ 0.134 \\
    Random - Relabel & 0.053 $\pm$ 0.006 \\
    \hline
    \end{tabular}
    \caption{Policy performance in the BabyAI environment trained on the augmented datasets from the different segmentation models. The best segmentation model is highlighted. All results are averaged over 8 seeds.}
    \label{tab:augmentation_results_babyai}
\end{table}

Since the play trajectories in the CALVIN environment have no annotations we cannot investigate the quality of the segmentations. To augment the starting split we follow \citet{labelling_model} and filter out labelled segments for which the labelling model has low confidence. The confidence threshold is set such that a label accuracy of $90\%$ on the validation set is achieved. In Table \ref{tab:augmentation_results_calvin} the results of the data augmentation are shown. Extracting labelled segments via video segmentation models improves policy performance. In comparison to the BabyAI environment a random segment from the play trajectories has a high likelihood of containing at most one instruction and additional frames from other instructions. Video segmentation models can then successfully crop away the undesirable frames as learned during training.

\begin{figure}[t!]
\centering
\includegraphics[width=\linewidth]{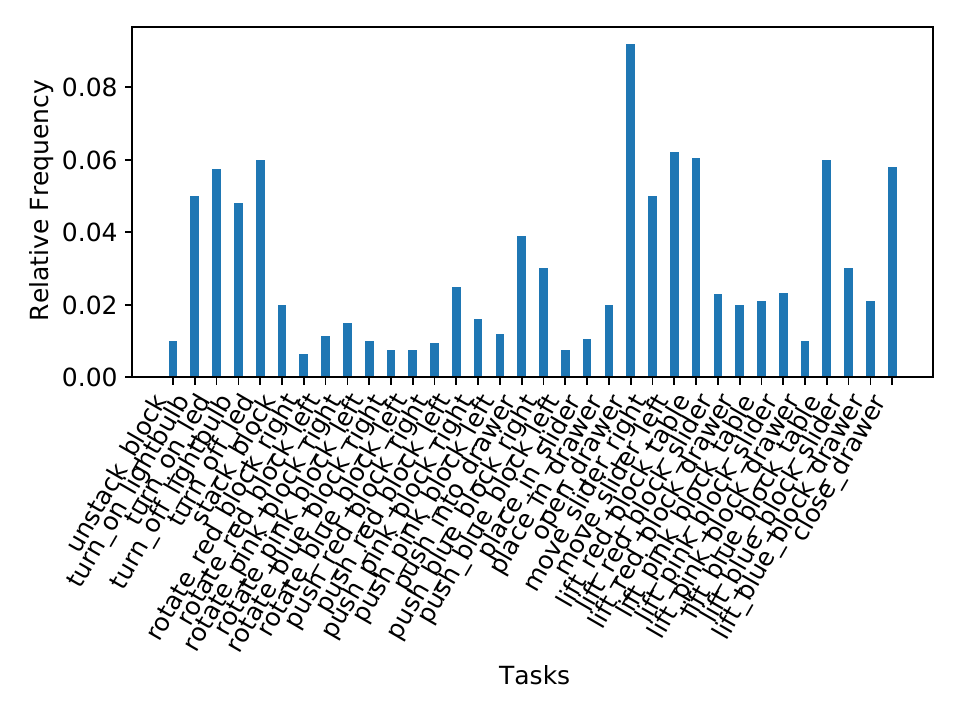}
\label{fig:sub1}
\caption{Distribution of labels in the segments extracted via Play Segmentation.}
\label{fig:label_distribution}
\end{figure}

\begin{figure}[t!]
    \centering
    \includegraphics[width=\linewidth]{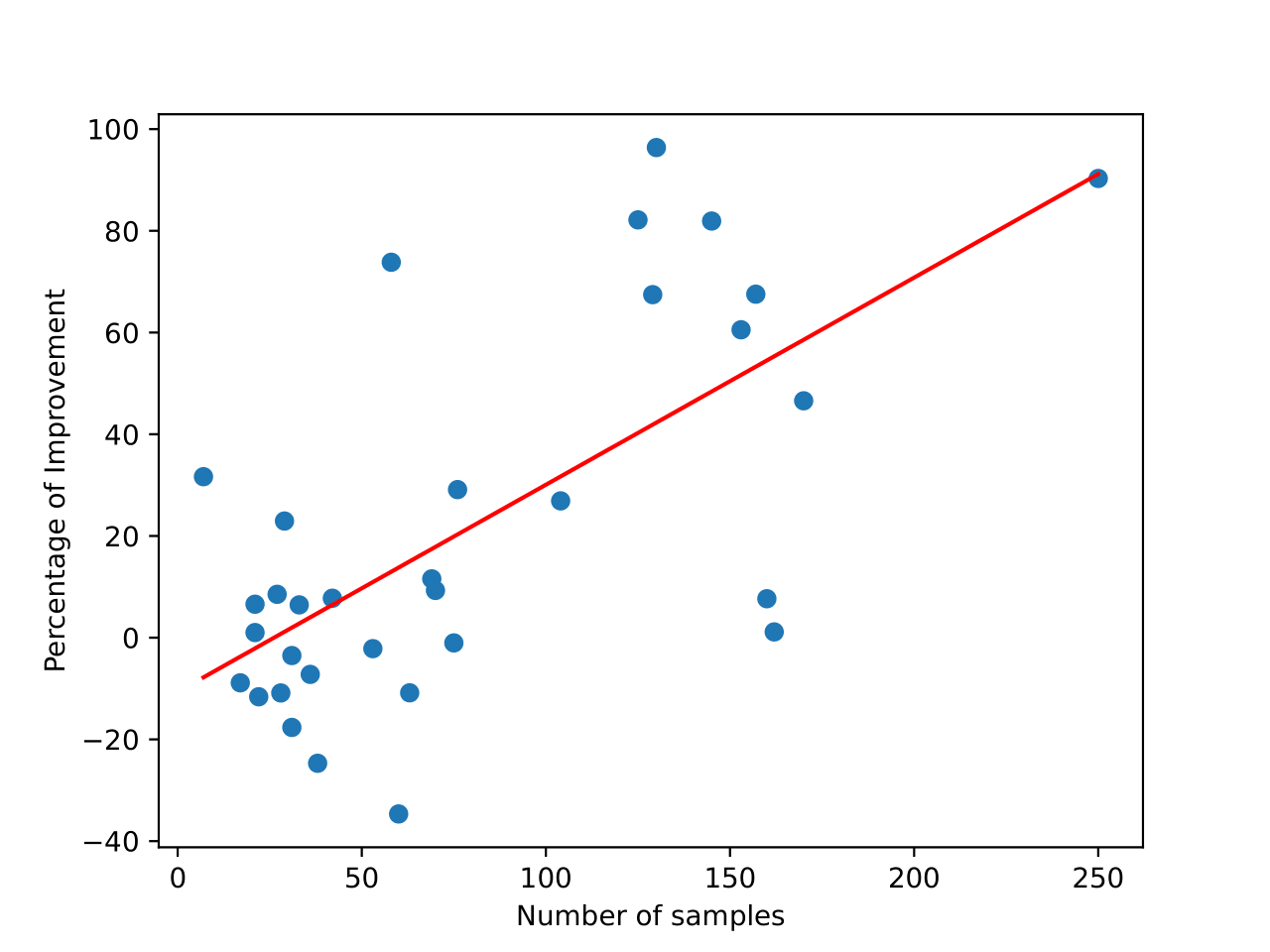}
    \caption{Displaying the relation between the number of samples added for a task and the improvement the policy achieves for that specific task in the CALVIN environment. The improvement is measure by the change in task accuracy between the original dataset and the augmented dataset relative to the improvement possible on that task.}
    \label{fig:correlation_n_samples_performance}
\end{figure}

Play segmentation outperforms the best video segmentation model and achieves a performance higher than if twice the labelled data would have been available. However, there is still a large gap to the relabelling groundtruth segments. To investigate the cause for the performance differences we looked at the label distribution of the extracted segments. In the original dataset the labels are uniformly distributed. From Figure \ref{fig:label_distribution} one can see that certain task labels are overrepresented in the added segments leading to an imbalanced dataset. There is a positive correlation between the amount of labelled segments added for a task and the policies improvement on the task (Figure \ref{fig:correlation_n_samples_performance}). The failure to extract labelled segments for certain tasks is a possible explanation for the observed performance differences.

\begin{table}[t]
    \centering
    \begin{tabular}{l | c }
    \hline
    Dataset & Avg. Seq. Length ($\pm$ Std.) \\
    \hline
    GT - 100\% & 1.698 $\pm$ 0.040 \\
    Relabel - GT & 1.679 $\pm$ 0.105\\
    PS  & \textbf{1.477 $\pm$ 0.092} \\
    GT - 50 \% & 1.411 $\pm$ 0.085 \\
    Tridet  & 1.394 $\pm$ 0.043\\
    Unloc & 1.315 $\pm$ 0.035 \\
    GT - 25\% & 1.230 $\pm$ 0.036\\
    Relabel - Random & 1.169 $\pm$ 0.084 \\
    GT - 10\%  & 0.867 $\pm$ 0.036 \\
    \hline
    \end{tabular}
    \caption{Performance of the MCIL policy in the CALVIN environment trained on the different augmented datasets. The best segmentation model is highlighted. All results are averaged over 4 seeds.}
    \label{tab:augmentation_results_calvin}
\end{table}

\section{Limitations}
Play Segmentation extracts better labelled segments than video segmentation models, which comes at the cost of higher computational complexity during segmentation. A solution could be to sample from the distribution over segmentations. This requires the learned distribution to only assign high probability to reasonable segmentations, otherwise segmentation errors will accumulate quickly. Another difficulty is to judge the quality of the augmented dataset if no groundtruth annotations are available. In realistic settings, such as the CALVIN dataset, the play trajectories have no annotation to check the quality of the extracted segments.

\section{Discussion}
We demonstrate the potential to leverage unsegmented trajectories for data augmentation to enhance an instruction-following policy in a play data setup. Play Segmentation shows better performance than adapted video segmentation methods. The possibility for training segmentation models depends on the type of annotated data available. The setup studied here has a special type of annotation, i.e. labelled subsegments of play trajectories. The tradeoff between the cost of different types of annotations and the possibilities for training annotation models is an area for future research.

\section*{Ethical Statement}
The long-term goal of the presented approach is to develop a text-to-trajectory model for game or robotic agents, similar to current text generation models. The two main risks are incorrect instruction execution and misuse by malicious actors. There is a trade-off between enhancing instruction-following policies and the potential for their misuse; better models can be more easily exploited. While the research aims to improve instruction following, positively impacting accuracy, it also heightens misuse risks.

\section{Acknowledgments}
This research was (partially) funded by the Hybrid Intelligence Center, a 10-year programme funded by the Dutch Ministry of Education, Culture and Science through the Netherlands Organisation for Scientific Research, https://hybrid-intelligence-centre.nl. This work used the Dutch national e-infrastructure with the support of the
SURF Cooperative using grant no. EINF-6630.

\bibliography{aaai25}
\newpage

\setcounter{secnumdepth}{1}
\appendix
\section{Details Segmentation Models}\label{app:seg_models}
\subsection{UnLoc}

\begin{figure}[h!]
    \centering
    \includegraphics[width=\linewidth]{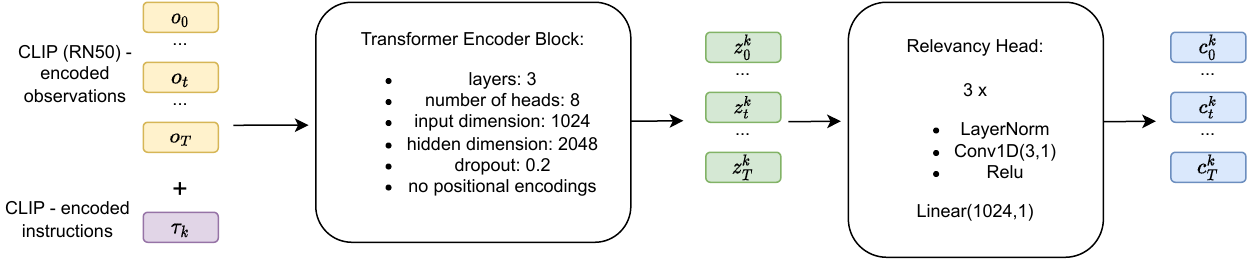}
    \caption{Overview of the UnLoc architecture. The number of layers N in the relevancy head depends on the dataset used for training (N=2 for BabyAI and N=3 for CALVIN). To get class probabilities for each frame, the displayed forward pass needs to happen for every instruction $\tau_{k}$ to then take the softmax over class-scores $c^{k}_{t}$.}
    \label{fig:unloc_arch}
\end{figure}

As no public implementation of UnLoc is available yet, we reimplement it. For an overview of the architecture and the training algorithm see the original work \cite{unloc}. The action segmentation version of UnLoc only takes into consideration part of the architecture. The feature pyramid and the boundary head are not necessary. We end up with the architecture shown in Figure \ref{fig:unloc_arch}. We augment the training segments from the annotated dataset by sampling random offsets to the left and right. The training details can be found in Table \ref{tab:training_details_unloc}. 

\textbf{Cropping Method}: At test time we sample a random window of the same size as used in training from the play trajectories and obtain frame-wise class labels via UnLoc. We then crop all contiguous background labels from the start and end of the segment. The resulting segment is only added to the labelled dataset if all of the class labels for the frames agree and the predicted class is not the background class. We also set a minimum size for added labelled segments, depending on the segment sizes that can be found in the training datasets.

\begin{table}[h!]
    \centering
    \begin{tabular}{l| c | c}
        Parameter & BabyAI & CALVIN \\
        \hline
        Learning Rate & 0.00005 & 0.00005\\
        Batch-Size & 128 & 24 \\
        $\#$ Updates & 20000 & 25000 \\
        Sampled Seg. Size & 25 & 100 \\
        Min. Seg. Size & 1 & 20 \\
        Hardware  & \makecell{NVIDIA \\ A100-SXM4- \\ 40GB} & \makecell{NVIDIA \\ TITAN X \\ (Pascal) 12GB} \\
        Wall-Clock Time & 10hrs. & 8hrs \\
    \end{tabular}
    \caption{Training details for UnLoc for the CALVIN environment and the BabyAI environment (Seg.=Segment).}
    \label{tab:training_details_unloc}
\end{table}

\subsection{TriDet}

We use the official implementation of TriDet \footnote{https://github.com/dingfengshi/TriDet}. The observations are encoded by taking the representation of the last layer of the labelling model (see Appendix  \ref{app:labelling_model}). If TriDet is trained on the $25\%$ data split, the corresponding labelling model is also trained on the same $25\%$ data split.
At training time we sample random offsets to the left and right of the segments from the annotated dataset to sample a fixed size random window.
The training details for TriDet can be found in Table \ref{tab:tridet_training_details}. For more details on the architecture and training loss we refer to the original implementation \cite{tridet}.

\begin{table}[]
    \centering
    \begin{tabular}{l | c | c}
      Parameter & BabyAI & CALVIN \\
      \hline 
      Learning Rate & 0.00005 & 0.00005 \\
      Batch-Size & 512 & 24\\
      $\#$ Updates & 2000 & 0.00005\\
      Min. Seg. Size & 1 & 20\\
      Sampled Seg. Size & 32 & 128 \\
      Hardware & \makecell{NVIDIA \\ A100-SXM4- \\ 40GB} & \makecell{NVIDIA GeForce \\ GTX 1080 Ti \\ 11GB} \\
      Wall-Clock Time & 1.5hr & 5hr. \\ 
    \end{tabular}
    \caption{Training Details for TriDet for the BabyAI and CALVIN environment (Seg.=Segment).}
    \label{tab:tridet_training_details}
\end{table}

\textbf{Cropping Method}:
At test time, we sample random windows from the play trajectories of the same size as during training. To extract a labelled segment from this window we take the frame of the feature pyramid with the highest classification confidence and take the associated predicted range for this frame as the corresponding segment and assign it the corresponding label. Segments that are below the minimum segment size are filtered out.

\subsection{Play Segmentation}

\begin{figure}[h!]
    \centering
    \includegraphics[width=\linewidth]{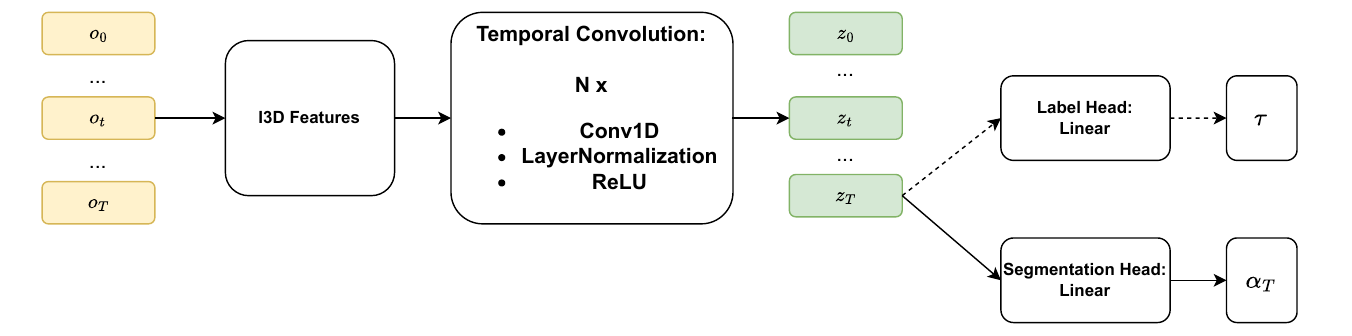}
    \caption{Overview of the Play Segmentation model. The number of temporal convolutions (N) depends on the dataset used for training (BabyAI N=2, CALVIN N=3).}
    \label{fig:play_segmentation_arch}
\end{figure}

The architecture of Play Segmentation is displayed in Figure \ref{fig:play_segmentation_arch} and the training details can be found in Table \ref{tab:training_details_play_segmentation}. Similarly to TriDet we first encode the environment observations using a pretrained labelling model (see Section \ref{app:labelling_model}).

\begin{table}[h!]
    \centering
    \begin{tabular}{l | c | c}
       Parameter & BabyAI & CALVIN \\
       \hline
       Learning Rate  & 0.00005 & 0.00005 \\
       Batch Size   & 128  & 24 \\
       $\#$ Updates   & 3000 & 12000 \\
       Min Seg. Size & 1 & 20 \\
       Max Seg. Size & 25 & 100 \\
       Hardware & \makecell{NVIDIA A100- \\
       SXM4-40GB} & \makecell{2 $\times$ NVIDIA \\ H100 90GB} \\
       Wall-Clock Time & 1hr & 14h 
       
    \end{tabular}
    \caption{The training details for Play Segmentation for the BabyAI environment and CALVIN environment (Seg.=Segment).}
    \label{tab:training_details_play_segmentation}
\end{table}

\textbf{Segmentation Algorithm:}
We adapt the recursion and algorithm presented in \citeauthor{sl3} to our setting. A key difference is that here the number of segments the trajectory should be segmented in, is not known a priori. We make use of the following recursion:

\begin{equation*} 
\begin{split}
    &\max_{\alpha_{0:T-1}} \log p_{\theta_{\textrm{Seg}}}(\alpha_{0:T-1}|o_{0:T}) = \\ 
    &\max_{i\in\{0,...,T-1\}} ( \max_{\alpha_{0:i}} \log p_{\theta_{\textrm{Seg}}}(\alpha_{0:i} | o_{0:i+1}) + \\
    & \log p_{\theta_{\textrm{Seg}}}(\alpha_{i+1:T-1}=(0,...,1)|o_{i+1:T})).
\end{split}
\end{equation*}
The recursion allows us to find the maximum likely segmentation up to timestep $t$ into $K$ segments by looking up for each timestep prior to $t$ the maximum likely segmentation into $K-1$ segments and the likelihood of the segment up to timestep $t$. To find the most likely segmentation, we find $\max_{\alpha_{0:T-1}} p_{\theta_{\textrm{Seg}}}(\alpha_{0:T-1}|o_{0:T})$ as described in Algorithm \ref{alg:DP} and keep track of the corresponding segmentation. The segmentation algorithm can be sped up by choosing a minimum and maximum window size for a segment. 

\begin{algorithm}[H]
\caption{Determining $\max \log p_{\theta_{\textrm{Seg}}}(\alpha_{0:T-1}|o_{0:T})$ }\label{alg:DP}
\begin{algorithmic}
\STATE {\bfseries Input:} $o_{0:T}$, $p_{\theta_{\textrm{Seg}}}$
\STATE 
Initialize $S_{ij}=-\infty$ for $i,j \in \{1,...,T\}$, where $S_{ij}$ is the log-likelihood of the optimal segmentation up to  timestep $j$ into $i$ intervals. Let $p^{\textrm{Seg}}_{ij} = p_{\theta_{\textrm{Seg}}}(\alpha_{j}=1|o_{i:j+1})$ and $S_{1k}= \sum^{k-1}_{t=0} \log (1-p^{\textrm{Seg}}_{0t}) + \log(p^{\textrm{Seg}}_{0k})$
\FOR{$i=2$ {\bfseries to} $T$}
    \FOR{$k=i$ {\bfseries to} $T$}
        \FOR{$l=i-1$ {\bfseries to} $k$}
        \STATE $S_{ik} = \max \{S_{ik},S_{i-1l} + \sum^{k-1}_{t=l+1}\log (1-p^{\textrm{seg}}_{lt}) + \log(p^{\textrm{seg}}_{lk})$ \}
        \ENDFOR
    \ENDFOR 
\ENDFOR
\STATE {\bfseries Return:} $\max_{i\in {1,...,T}} S_{iT}$

\end{algorithmic}
\end{algorithm}

\section{Details Policy Training}\label{app:policy_training}

\subsection{Calvin}
We train a policy via multi-context imitation learning (MCIL) \cite{mcil_original,calvin,hulk}. We follow the implementation that was released alongside the CALVIN benchmark \footnote{https://github.com/mees/calvin}. The RNN policy block is replaced by the encoder block of a Transformer \cite{transformer} with 6 layers and a context window of 4. Apart from the learning rate hyperparameters are unchanged from the values found in the implementation. The training parameters can be found in Table \ref{tab:training_details_policy}. 

\subsection{BabyAI}

The policy implementation follows \citet{sample_efficiency_babyai}, but we do not embed the instructions via a GRU \cite{gru} but feed pretrained sentence embeddings directly to the FILM layer \cite{film_layer}. The pretrained sentence embeddings are obtained from the last hidden layer of a variant of the T5 encoder \cite{t5} \footnote{https://huggingface.co/google-t5/t5-small}. Further, we remove the LSTM component from the policy. The BabyAI environment used here is fully observable. The training parameters are listed in Table \ref{tab:training_details_policy}. Hyperparameters are tuned manually and chosen based on the evaluation performance on the full annotated dataset. 

\begin{table}[h!]
    \centering
    \begin{tabular}{ l | c | c }
     Parameter & BabyAI & CALVIN \\
     \hline
     Learning Rate & 0.0001 & 0.00005 \\
     Batch Size    &  256 & 64 \\
     $\#$ Updates & 20000 & 45000 \\
     Hardware  &  \makecell{GeForce GTX \\ 1080 Ti (11GB)} & \makecell{NVIDIA \\ A100-SXM4-40GB}\\ 
     Wall-Clock T. &  20min. & 18hrs. \\
    \end{tabular}
    \caption{The training parameters for training an imitation learning policy in the BabyAI and CALVIN environment.}
    \label{tab:training_details_policy}
\end{table}

\section{Details Labelling Model}\label{app:labelling_model}

We train a video classifier following the I3D architecture \cite{i3d} on the annotated dataset consisting of instruction-trajectory pairs. We use the implementation from PySlowFast \footnote{https://github.com/facebookresearch/SlowFast}. The training details are displayed in Table \ref{tab:training_details_lm}. Hyperparameters are tuned manually until a satisfactory validation set performance is achieved.

\begin{table}[h!]
    \centering
    \begin{tabular}{l | c | c}
      Parameter & BabyAI & CALVIN \\
      \hline
       Learning Rate  &  0.0001 & 0.00005 \\
       Batch-Size & 256 & 48 \\
       $\#$ Updates & 8000 & 32000 \\
       Hardware & A100 GPU & 4 $\times$ A100 GPU \\
       Wall-Clock Time & 1hr. & 6.5hrs.
       
    \end{tabular}
    \caption{Training details for the I3D video classifier in the BabyAI and CALVIN environment. The NVIDIA A100 GPUs have 40GB of memory.}
    \label{tab:training_details_lm}
\end{table}

\end{document}